\documentclass{article}

\usepackage{microtype}
\usepackage{graphicx}
\usepackage{subcaption}
\usepackage{threeparttable}
\usepackage{booktabs} 
\usepackage{algpseudocode}
\usepackage{array}
\usepackage{booktabs}
\usepackage{xcolor}
\usepackage{rotating}
\usepackage{hyperref}

\usepackage[preprint]{icml2026}

\usepackage{amsmath}
\usepackage{amssymb}
\usepackage{mathtools}
\usepackage{amsthm}

\usepackage[capitalize,noabbrev]{cleveref}

\definecolor{gred}{RGB}{200, 50, 50}
\definecolor{ggreen}{RGB}{50, 150, 50}
\definecolor{gblue}{RGB}{50, 50, 150}
\theoremstyle{plain}

\theoremstyle{definition}

\theoremstyle{remark}

\usepackage[textsize=tiny]{todonotes}

\icmltitlerunning{QVAD: Question-Centric Agentic Framework for Video Anomaly Detection}

\begin{document}

\twocolumn[
  \icmltitle{QVAD: A Question-Centric Agentic Framework for\\
    Efficient and Training-Free Video Anomaly Detection}


  \icmlsetsymbol{equal}{*}

\begin{icmlauthorlist}
    \icmlauthor{Lokman Bekit}{univ}
    \icmlauthor{Hamza Karim}{univ}
    \icmlauthor{Nghia T Nguyen}{univ}
    \icmlauthor{Yasin Yilmaz}{univ}
\end{icmlauthorlist}

\icmlaffiliation{univ}{Department of Electrical Engineering, University of South Florida, Tampa, Florida, USA}

\icmlcorrespondingauthor{Lokman Bekit}{lbekit@usf.edu}
\icmlcorrespondingauthor{Hamza Karim}{hamzakarim@usf.edu}
\icmlcorrespondingauthor{Nghia T Nguyen}{nghianguyen@usf.edu}
\icmlcorrespondingauthor{Yasin Yilmaz}{yasiny@usf.edu}
  \icmlkeywords{Video Anomaly Detection, Vision-Language Models, Large Language Models, Agentic AI, Zero-Shot Learning}

  \vskip 0.3in
]


\printAffiliationsAndNotice{}  

\begin{abstract} Video Anomaly Detection (VAD) is a fundamental challenge in computer vision, particularly due to the open-set nature of anomalies. While recent training-free approaches utilizing Vision-Language Models (VLMs) have shown promise, they typically rely on massive, resource-intensive foundation models to compensate for the ambiguity of static prompts. We argue that the bottleneck in VAD is not necessarily model capacity, but rather the static nature of inquiry. We propose QVAD, a question-centric agentic framework that treats VLM-LLM interaction as a dynamic dialogue. By iteratively refining queries based on visual context, our LLM agent guides smaller VLMs to produce high-fidelity captions and precise semantic reasoning without parameter updates. This ``prompt-updating" mechanism effectively unlocks the latent capabilities of lightweight models, enabling state-of-the-art performance on UCF-Crime, XD-Violence, and UBNormal using a fraction of the parameters required by competing methods. We further demonstrate exceptional generalizability on the single-scene ComplexVAD dataset. Crucially, QVAD achieves high inference speeds with minimal memory footprints, making advanced VAD capabilities deployable on resource-constrained edge devices. \end{abstract}

\section{Introduction}Video Anomaly Detection (VAD) is a critical component of intelligent surveillance systems, yet it remains a persistent challenge due to the unbounded and open-set nature of what constitutes an ``anomaly.'' Traditional specialist models, while efficient, are brittle; they rely on extensive domain-specific supervision and fail to generalize to new environments without expensive retraining. To overcome this, recent paradigms have shifted toward training-free methods leveraging Large Language Models (LLMs) and Vision-Language Models (VLMs). 

While effective, these frameworks typically suffer from a dual burden: they rely on static prompts (e.g., ``Is there an anomaly?'') and, consequently, require massive parameter counts to interpret complex temporal dynamics correctly. In standard approaches like LAVAD~\cite{zanella2024lavad}, the lack of prompt specificity forces the reliance on heavy, distinct foundation models to handle ambiguity. This 
confines training-free VAD to server-grade hardware, disallowing edge deployment. 

We propose that the key to practical VAD lies not in scaling up model size, but in scaling up the quality of interaction. We introduce QVAD, a Question-Centric Agentic Framework that bridges the performance gap for lightweight models through inference-time optimization. Our core hypothesis is that smaller VLMs yield significantly better visual reasoning when the prompt is structurally optimized. Instead of a one-shot query, QVAD treats the VAD process as a dynamic dialogue and can be interpreted as an instance of active multimodal inference, where the LLM selects questions as actions to reduce uncertainty over anomaly hypotheses. Unlike prior static prompting approaches, QVAD implements a closed-loop perception system. 
This iterative prompt updating allows a $2\text{--}4$ billion parameter model to extract visual evidence with a level of nuance typically reserved for larger models. 

We validate QVAD on major benchmarks including UCF-Crime, XD-Violence, and UBNormal, achieving state-of-the-art results. Furthermore, we evaluate our framework on the challenging ComplexVAD~\cite{mumcu2025complexvad} dataset, demonstrating successful generalization to the specific context-dependent anomalies of a single-scene environment. Most importantly, our ablation studies confirm that QVAD democratizes access to advanced VAD. By achieving high accuracy with compact models (e.g., Qwen-1.7B), our framework operates within a memory footprint as low as $4.20$ GB. This efficiency confirms that QVAD is not only accurate but uniquely viable for real-world deployment on edge computing platforms like the NVIDIA Jetson Orin Nano. 
Our main contributions are as follows:
\begin{itemize}
    \item \textbf{Dynamic Question Refinement:} We introduce an inference-time mechanism in which an LLM-based agent adaptively updates VLM prompts on-the-fly, maximizing information retrieval in ambiguous or uncertain scenes.
    
    \item \textbf{Efficient Deployment:} We show that the proposed agentic loop maintains strong performance when paired with smaller foundation models, effectively democratizing access to state-of-the-art video anomaly detection (VAD) capabilities.

    \item \textbf{Generalization to Open-Set Scenarios:} We demonstrate that QVAD successfully generalizes to the context-dependent characteristics of a practical single-scene environment. 
\end{itemize}

\section{Related Work}

The field of Video Anomaly Detection has evolved from reconstruction-based deep learning methods to recent training-free approaches leveraging multimodal foundation models. This section reviews the progression of these methodologies, focusing on prompt optimization, structural video understanding, open world methods, and agentic frameworks.

\subsection{Training-Dependent VAD}

Traditional Video Anomaly Detection (VAD) predominantly relies on reconstruction errors or discriminative learning, operating with varying degrees of supervision—ranging from weakly-supervised approaches \cite{tian2021rtfm, karim2024real, chen2024prompt, wu2024vadclip, wu2020not} to fully unsupervised methods \cite{qi2025time, zhang2024multi, al2024collaborative, ristea2024self}. More recent efforts have begun to explore semi-supervised and hybrid SSL–LLM paradigms to mitigate this limitation, leveraging multimodal semantic descriptions and language-guided reasoning to improve interpretability and robustness in complex anomaly scenarios \cite{mumcu2025leveraging, hemmatyar2025hycovad}. While these methods achieve high performance within closed domains, they suffer from poor generalization to unseen environments.

To address this, recent weakly-supervised methods have begun integrating Vision-Language Models (VLMs). For example, VadCLIP~\cite{wu2024vadclip} adapts a frozen CLIP model by introducing learnable temporal adapters and prompt vectors. However, despite leveraging VLM features, VadCLIP remains fundamentally training-dependent; it requires domain-specific training data to align visual and textual modalities, thereby restricting its zero-shot transferability to novel anomaly types.

\subsection{From Offline Optimization to Inference-Time Reasoning}

The shift towards training-free VAD aims to leverage the native zero-shot capabilities of VLMs (e.g., CLIP, InternVL) without requiring parameter updates or domain-specific data. Early attempts, such as ~\cite{zanella2024lavad, shao2025eventvad} utilized frozen VLMs with static captions. However, these methods often fail to capture subtle temporal anomalies due to the ``semantic gap'' between generic pre-training data and specific anomaly definitions.

To bridge this gap, recent frameworks have integrated complex reasoning chains. However, distinct from our approach, these methods effectively reintroduce a dependency on training data. For instance, VERA~\cite{ye2025vera} proposes ``Verbalized Learning,'' a framework that decomposes reasoning into guiding questions. Crucially, VERA is not training-free; it optimizes these questions \textit{offline} using a learner-optimizer loop driven by coarse video-level labels from a training set. 

Similarly, while Holmes-VAD~\cite{zhang2025holmes} utilizes Multi-Modal Large Language Models (MLLMs) for granular analysis, it relies on a multi-stage supervised training pipeline. Specifically, Holmes-VAD trains a dedicated anomaly scorer using annotated frame-level labels (e.g., from HIVAU-70k) and subsequently fine-tunes the MLLM via LoRA to align it with instruction data. This reliance on supervised fine-tuning and large-scale annotated datasets fundamentally separates it from zero-shot paradigms.

In contrast, our framework eliminates the need for both parameter updates (as in Holmes-VAD) and offline prompt optimization (as in VERA). We employ   inference-time refinement, generating an adaptive questioning strategy dynamically. This allows our model to construct the most relevant questions for each specific video instance on the fly, akin to an active sensing strategy, without ever seeing the source distribution or requiring annotated supervision.

\begin{figure*}[t]
    \centering
    \includegraphics[width=\textwidth]{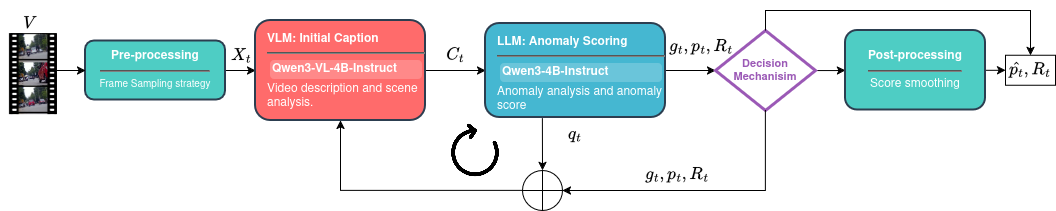}
    \caption{Overview of the proposed QVAD framework. The VLM and LLM agents engage in an iterative dialogue, where the LLM generates a query $q_t$ conditioned on the captions $C_t$ produced by the VLM and directs it back to the VLM. This feedback loop progressively refines their shared understanding of the scene}
    \label{fig:Simplified_figure}
    \vspace{-3mm}
\end{figure*}

\subsection{Open-Vocabulary and Generalist VAD}

Research has increasingly prioritized generalizability to handle open-set anomalies. Open-Vocabulary Video Anomaly Detection (OVVAD)~\cite{wu2024ovvad} addresses the ``unseen anomaly'' challenge via a two-stage pipeline: detecting generic anomalies first, then classifying them using semantic knowledge. While effective, OVVAD treats detection and semantic classification as decoupled tasks. Our framework unifies these processes through integrated natural language dialogue, allowing detection to be directly guided by semantic reasoning.

Similarly, ``Follow the Rules''~\cite{yang2024rules} explores agentic reasoning by translating normality definitions into textual rules for an LLM verification. However, this approach relies on static rule sets. Our work advances this paradigm by making the ``rules'', formulated as guiding questions, dynamic; our agent iteratively evolves its inquiry based on the unfolding visual context, a capability essential for identifying context-dependent irregularities.

\subsection{Structural and Agentic Video Understanding}

Two recent methods address the complexity of long-term video dependencies. VADTree~\cite{li2025vadtree} focuses on structural reasoning, proposing a Hierarchical Granularity-aware Tree to organize video segments into clusters for multi-scale analysis. 
While VADTree optimizes temporal structure, PANDA~\cite{yang2025panda} explores agentic structure, deploying a comprehensive ``AI Engineer'' capable of planning, tool use (e.g., web search), and self-reflection. However, PANDA achieves performance at the cost of significant system complexity and latency. Our work strikes a critical balance: we adopt an agentic stance where the LLM actively reasons and plans, but we constrain the action space to iterative question refinement. This yields the adaptability of complex agents like PANDA while maintaining the efficiency required for practical deployment.

\section{Method}

We propose a \emph{Question-centric agentic framework} for Video Anomaly Detection (QVAD). Unlike traditional methods that rely on static visual features or fixed text prompts, our approach models anomaly detection as a dynamic, multi-turn dialogue between a visual perception module (VLM) and a reasoning agent (LLM) as shown in Figure \ref{fig:Simplified_figure}.\par

\begin{figure*}[h]
    \centering
    \includegraphics[width=\textwidth]{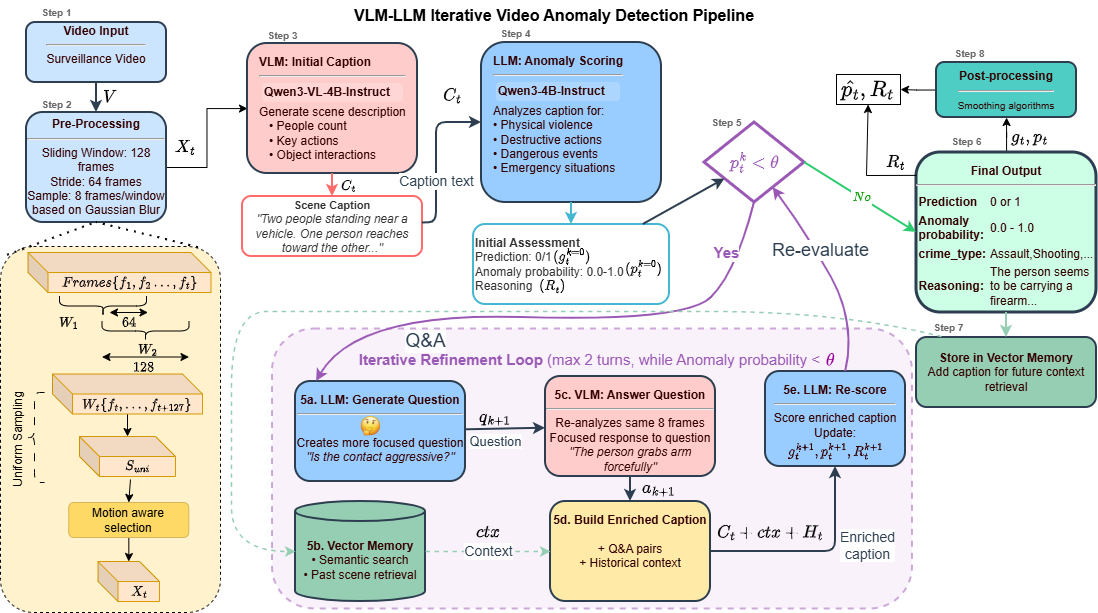}
    \caption{Detailed architecture of the proposed QVAD framework.}
    \label{fig:Detailed_figure}
    \vspace{-3mm}
\end{figure*}

Figure \ref{fig:Detailed_figure} shows a more detailed illustration of the QVAD architecture. Let a video sequence be denoted by $\mathcal{V} = \{f_1, f_2, \dots, f_T\}$, where $f_i$ represents the $i$-th frame. We process the video through a pre-processing step using a sliding window approach with a window size of $w=128$ frames and a stride of $s=64$ frames. For a given temporal window $\mathcal{W}_t = \{f_t, \dots, f_{t+127}\}$, our goal is to derive an anomaly score $y_t \in [0, 1]$ and a corresponding semantic explanation.

\subsection{Two-Stage Spatiotemporal Abstraction}

To efficiently capture both long-term temporal context and fine-grained motion dynamics within the window $\mathcal{W}_t$, we employ a cascaded frame selection strategy. Direct processing of all 128 frames in $\mathcal{W}_t$ is computationally prohibitive for VLMs; however, random or purely uniform sampling risks missing transient anomalies (e.g., a sudden snatch-and-grab). Algorithm \ref{alg:motion} in Appendix and Figure \ref{fig:Detailed_figure} detail the extraction of relevant frames in the pre-processing stage.

\textbf{Stage 1: Uniform Temporal Coverage.} First, we convert the image to grayscale and reduce the redundancy of high-FPS video by uniformly sampling a set of candidate frames $\mathcal{S}_{uni}$ to ensure temporal coverage across the entire window:
\begin{equation}
    \mathcal{S}_{uni} = \{f_{t + i \cdot \Delta} : i \in \{0, \dots, N_{uni}-1\} \}
\end{equation}
where $N_{uni} = 32$ is the budget for the uniform sampler and $\Delta=w/N_{uni}$. This reduces the input from 128 frames to a manageable 32 frames while maintaining the temporal breadth of the window.

\textbf{Stage 2: Motion-Aware Salience Selection.} From $\mathcal{S}_{uni}$, we select a final subset $\mathcal{S}_{motion}$ of $N_{vlm}=8$ frames that exhibit the highest salient motion. We define the motion saliency $m_i$ using the mean intensity of the Gaussian-blurred difference between consecutive frames:
\begin{equation}
    m_i = \frac{1}{H \times W} \sum_{x,y} | G_\sigma(f_i(x,y)) - G_\sigma(f_{i+1}(x,y)) |
\end{equation}
where $H$ and $W$ are the height and width of the frame, $x,y$ are pixel coordinates and $G_\sigma$ denotes a Gaussian smoothing filter. We retain the first and last frames of $\mathcal{S}_{uni}$ to preserve boundary context and select the remaining $6$ frames with the highest $m_i$ scores. This yields the final visual input tensor $\mathbf{X}_t = \mathcal{S}_{motion}$ (where $|\mathbf{X}_t|=8$), ensuring the VLM attends to the most dynamic segments of the window.

\subsection{Iterative Agentic Reasoning}

The core of our framework is an iterative dialogue loop where an LLM Agent ($\mathcal{A}$) refines its understanding of the visual input $\mathbf{X}_t$ by querying a VLM Perception module ($\mathcal{P}$).

\textbf{Initialization.} At turn $k=0$, the perception module generates an initial comprehensive caption $C_t$ based on a broad system prompt $P_{init}$ (e.g., ``Describe the persons, actions, and interactions''):
\begin{equation}
    C_{t} = \mathcal{P}(\mathbf{X}_{t}, P_{init})
\end{equation}

\textbf{The Reasoning Loop.} The LLM agent first assesses the probability of an anomaly based on the current context. The agent outputs a tuple $(g_t^k, p_t^k, R_t^k) = \mathcal{A}(C_{t}, \mathcal{H}_{t}, \text{ctx})$, $k=0,1,\ldots$, where $g_t^k \in \{0,1\}$ is the binary anomaly prediction, $p_t^k$ is the anomaly probability, $R_t^k$ is the reasoning, $\mathcal{H}_t=\{(q_t^k,a_t^k): k=1,\ldots,K\}$ is the conversation history from all turns, consisting of the accumulated question-answer pairs, and  \text{ctx} is the context retrieved from vector memory (Eq. \eqref{eq:ctx}).

If the probability $p_t^k$ is below a threshold $\theta$, the agent initiates the next turn by formulating a clarifying question $q_t^{k+1}$ to resolve ambiguity. Crucially, this question is conditioned on the previous dialogue and the specific ambiguity detected in $R_t^k$:
\begin{equation}
    q_t^{k+1} = \mathcal{A}(C_t, \mathcal{H}_t, R^k_t,p^k_t,g^k_t).
\end{equation}

The VLM then provides a targeted answer $a_t^{k+1}$, attending specifically to the visual regions relevant to $q_t^{k+1}$ based on a prompt $(P_{focus})$ designed to respond to the question $q_{k+1}$:
\begin{equation}
    a_t^{k+1} = \mathcal{P}(\mathbf{X}_t, q_t^{k+1},P_{focus})
\end{equation}

The history is updated as $\mathcal{H}_t = \mathcal{H}_t \cup (q_t^{k+1}, a_t^{k+1})$, and this process repeats until $p_t^k>\theta=0.7$ or a maximum turn limit $K$ is reached. This adaptive mechanism allows the model to distinguish hard negatives (e.g., ``running for exercise'' vs. ``fleeing'') by actively seeking discriminative visual details. The reasoning algorithm is detailed in Algorithm~\ref{alg:main_pipeline} in Appendix.

\subsection{Vector Memory Module}
We implement a Vector Memory module that couples a dense retrieval index $\mathcal{I}$ with a metadata storage $\mathbf{M}_v$. The mechanisms for storage and retrieval are formally defined in Algorithm~\ref{alg:memory} (Appendix).

During the storage phase, the agent's current observation $C_t$, anomaly prediction $g^k_t$, and immediate history $\mathcal{H}_t$ are concatenated and truncated to a fixed capacity $\tau_{\text{cap}}$. This text is mapped to a latent vector $\mathbf{e} \in \mathbb{R}^{384}$ via a pre-trained encoder $\mathbf{E}$ and committed to the index. To maintain computational efficiency, the index undergoes a pruning process whenever the storage size exceeds a threshold of $\tau_{\max}$.

Retrieval is performed via semantic search in the latent space. Given a query, we generate an embedding $\mathbf{q}_e$ using $\mathbf{E}$ and identify the top-$K$ nearest neighbors within $\mathcal{I}$ based on cosine similarity to form the context used by the LLM agent for anomaly predictions: 
\begin{equation}
    \text{ctx} = \left\{ \mathbf{M_v}(j) : j \in \underset{i\in\mathcal{I}}{\text{arg top-}K} \cos(\mathbf{q}_e, \mathbf{e}_i) \right\}.
    \label{eq:ctx}
\end{equation}
Rather than returning a fixed number of documents, the system aggregates metadata from the ranked results dynamically. This accumulation terminates once the total length of retrieved segments reaches a token budget $\tau_{\text{ctx}}$, ensuring the agent receives maximal relevant context without exceeding the prompt window.





\subsection{Post Processing}
\paragraph{Frame-level anomaly score generation.}
For each window $\mathcal{W}_t$, the model outputs a binary prediction $g_t \in \{0,1\}$ and an anomaly probability $p_t \in [0,1]$ after last turn. 
Since consecutive windows overlap, each frame $f_i$ can belong to multiple windows. Let
\[
\mathcal{T}(i) = \{ t : f_i \in \mathcal{W}_t \}
\]
denote the set of all windows covering frame $f_i$. We aggregate the window-level outputs into frame-level predictions by max pooling over the overlapping windows:
\[
g_i = \max_{t \in \mathcal{T}(i)} g_t, 
\qquad 
p_i = \max_{t \in \mathcal{T}(i)} p_t.
\]

This aggregation preserves the strongest anomaly evidence in overlapping regions, ensuring that a frame is considered anomalous if any of its associated temporal contexts indicates high anomaly confidence. We then apply the following weighting strategy to compute a calibrated frame-level score
$L_i$ as

\begin{equation}
L_i =
\begin{cases}
\max(p_i, \alpha), & \text{if } g_i = 1, \\
\min(p_i, \alpha), & \text{if } g_i = 0,
\end{cases}
\label{eq:superadd}
\end{equation}

where $\alpha$ is a hyper-parameter that enforces a lower bound on the confidence
for anomalous clips and an upper bound on the anomaly score for normal clips.
This prevents suppression or amplification of anomaly scores.

Next, we apply a multi-scale Gaussian smoothing operation to capture temporal
consistency. Let $\mathcal{G}_{\sigma_1}(\cdot)$ denote a Gaussian convolution
with variance $\sigma_1$. The smoothed frame-level scores are given by

\begin{equation}
\tilde{L}_i = \mathcal{G}_{\sigma_1}(L_i).
\label{eq:gaussian1}
\end{equation}

Finally, we refine the scores using the frame-level anomaly scoring method
proposed in Vera \textit{et al.}, which further enforces temporal smoothness and
robustness via another Gaussian kernel with variance $\sigma_2$. Denoting this
operation by $\mathcal{V}_{\sigma_2}(\cdot)$, the final frame-level anomaly score
is computed as

\begin{equation}
\hat{p}_i = \mathcal{V}_{\sigma_2}(\tilde{L}_i).
\label{eq:vera}
\end{equation}




\section{Experiments}

\subsection{Experimental Setup}
\textbf{Datasets.} We evaluate QVAD on four diverse benchmarks to assess generalizability across real-world, synthetic, and long-form video scenarios. UCF-Crime ~\cite{sultani2018real} and XD-Violence~\cite{wu2020not}  serve as standard benchmarks for real-world surveillance, containing anomalies such as abuse, fighting, and theft. To test open-set adaptability, we utilize UBNormal ~\cite{acsintoae2022ubnormal} and ComplexVAD ~\cite{mumcu2025complexvad}, a recently introduced single-scene dataset that challenges models with long-temporal contexts and subtle, non-binary anomalies.

\textbf{Metrics.} Following standard protocols, we report the Receiver Operating Characteristic Area Under the Curve (AUC) for frame-level anomaly detection on UCF-Crime, UBNormal, and ComplexVAD. For XD-Violence, we report Average Precision (AP), consistent with prior literature.

\textbf{Implementation.} All evaluations were conducted on a single NVIDIA RTX 5090 GPU. Our primary results are obtained using compact models. Our default configuration pairs a standard Qwen3 4B LLM agent with a Qwen3 4B VLM \cite{bai2025qwen3vl, yang2025qwen3}, resulting in a total parameter count of approximately 8B. Optimal values for $\sigma_1$, $\sigma_2$ and $\alpha$ for each dataset are given in Table \ref{tab:post-process hyperparameters} in the appendix.

To rigorously evaluate deployment feasibility, we test two inference strategies: 
1. \textbf{Standard Pipeline:} Both the LLM and VLM are loaded simultaneously into VRAM to maximize throughput. 
2. \textbf{Memory-Optimized Pipeline:} Models are sequentially loaded and offloaded (VLM $\to$ Inference $\to$ Offload $\to$ LLM), minimizing peak memory usage at the cost of latency.

\subsection{Results}
\begin{table}[t]
\caption{Performance comparison on UCF-Crime and XD-Violence benchmarks. We report AUC (\%) for UCF-Crime and AP (\%) for XD-Violence. ``Expl.'' indicates explainability.}
\label{tab:vad_comparison1}
\vskip 0.15in
\begin{center}
\begin{small}
\begin{sc}
\setlength{\tabcolsep}{3pt}
\begin{tabular}{llccc}
\toprule
Methods & Supervision & Expl. & UCF & XD \\
       &             &       & (AUC) & (AP) \\
\midrule
\multicolumn{5}{l}{\textit{Specialized}} \\
HL-Net         & Weak    & $\times$     & 82.44 & 73.67 \\
RTFM         & Weak    & $\times$     & 84.30 & 77.81 \\
VadCLIP       & Weak    & $\times$     & 88.02 & 84.51 \\
VERA          & Weak    & $\checkmark$ & 86.55 & 70.11 \\
Holmes-VAU    & Instru. & $\checkmark$ & 88.96 & 87.68 \\
\midrule
\multicolumn{5}{l}{\textit{Training-free}} \\
LAVAD         & Train-free & $\checkmark$ & 80.28 & 62.01 \\
EventVAD      & Train-free & $\checkmark$ & 82.03 & 64.04 \\
PANDA         & Train-free & $\checkmark$ & 84.89 & 70.16 \\
VADTree       & Train-free & $\checkmark$ & 84.74 & 68.85 \\
\midrule
\textbf{QVAD} & Train-free & $\checkmark$ & 84.28 & 68.53 \\
\bottomrule
\end{tabular}
\end{sc}
\end{small}
\end{center}
\vskip -0.1in
\end{table}

\begin{table}[t]
\caption{Performance comparison on UBNormal (UB) and ComplexVAD (CVAD) benchmarks. We report AUC (\%) for both datasets.}
\label{tab:vad_comparison2}
\vskip 0.15in
\begin{center}
\begin{small}
\begin{sc}
\setlength{\tabcolsep}{3pt}
\begin{tabular}{llccc}
\toprule
Methods & Supervision & Expl. & UB & CVAD \\
       &             &       & (AUC) & (AUC) \\
\midrule
\multicolumn{5}{l}{\textit{Specialized}} \\
RTFM          & Weak    & $\times$   & 64.94  & --    \\
HyCoVAD       & SSL    & $\checkmark$    & --    & 72.5   \\
MLLM-EVAD     & Semi    &   $\checkmark$ & --     & 71.0 \\
Holmes-VAU    & Instru. & $\checkmark$ & 56.77 & -- \\
\midrule
\multicolumn{5}{l}{\textit{Training-free}} \\
LAVAD         & Train-free & $\checkmark$ & 64.23 & --    \\
Follow the Rules  & Train-free & $\checkmark$ & 71.90 & --    \\
PANDA         & Train-free & $\checkmark$ & 75.78 & --    \\
VADTree      & Train-free & $\checkmark$ & 65.80 & 54.20 \\
\midrule
\textbf{QVAD} & Train-free & $\checkmark$ & 79.6 & 68.02 \\
\bottomrule
\end{tabular}
\end{sc}
\end{small}
\end{center}
\vskip -0.1in
\end{table}
\begin{figure*}[t]
\centering
\footnotesize                    
\renewcommand{\arraystretch}{0.95}
\setlength{\tabcolsep}{3pt}      

\begin{tabular}{p{0.02\textwidth}
p{0.15\textwidth}
p{0.28\textwidth}
p{0.50\textwidth}}
\toprule
& \textbf{Input Scene} & \textbf{Turn 0: Static Baseline (Failure)} & \textbf{Turn 1: QVAD Agentic Refinement (Success)} \\
\midrule

\raisebox{-1.2cm}[0pt][0pt]{\rotatebox{90}{\textbf{Arrest07}}} &
\begin{minipage}[t]{\linewidth}\vspace{0pt}\centering
    \includegraphics[width=\linewidth,height=2.2cm,keepaspectratio]{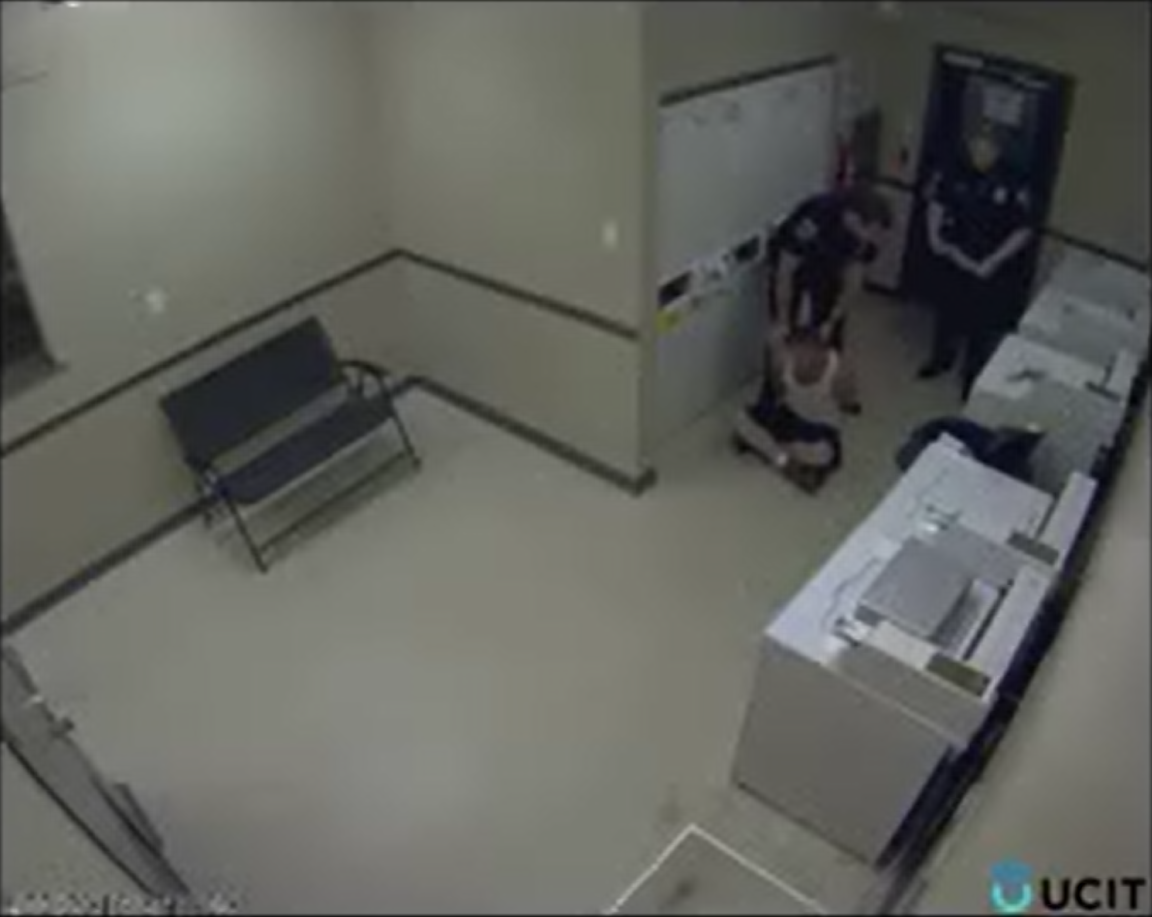}
\end{minipage}
&
\begin{minipage}[t]{\linewidth}\vspace{0pt}
    \setlength{\parskip}{0pt}
    \setlength{\parindent}{0pt}
    \linespread{0.95}\selectfont

    \textbf{VLM Caption:} ``One person sitting on floor... two others standing over them.'' \par
    \textbf{LLM Reasoning:} ``No explicit indication of violence... likely a potential escalation but no immediate danger.'' \par
    \textcolor{red}{\textbf{Prediction: Normal}}
\end{minipage}
&
\begin{minipage}[t]{\linewidth}\vspace{0pt}
    \setlength{\parskip}{0pt}
    \setlength{\parindent}{0pt}
    \linespread{0.95}\selectfont

    \textbf{Agent Query (T1):} \textit{``Is the seated person being physically restrained or threatened?''} \par
    \textbf{VLM Response:} ``Yes, the seated person is being physically restrained.'' \par
    \textbf{LLM Re-eval:} ``Physical restraint meets criteria for Assault.'' \par
    \textcolor{blue}{\textbf{Prediction: Anomalous}}
\end{minipage}
\\
\midrule

\raisebox{-1.6cm}[0pt][0pt]{\rotatebox{90}{\textbf{Stealing019}}} &
\begin{minipage}[t]{\linewidth}\vspace{0pt}\centering
    \includegraphics[width=\linewidth,height=2.5cm,keepaspectratio]{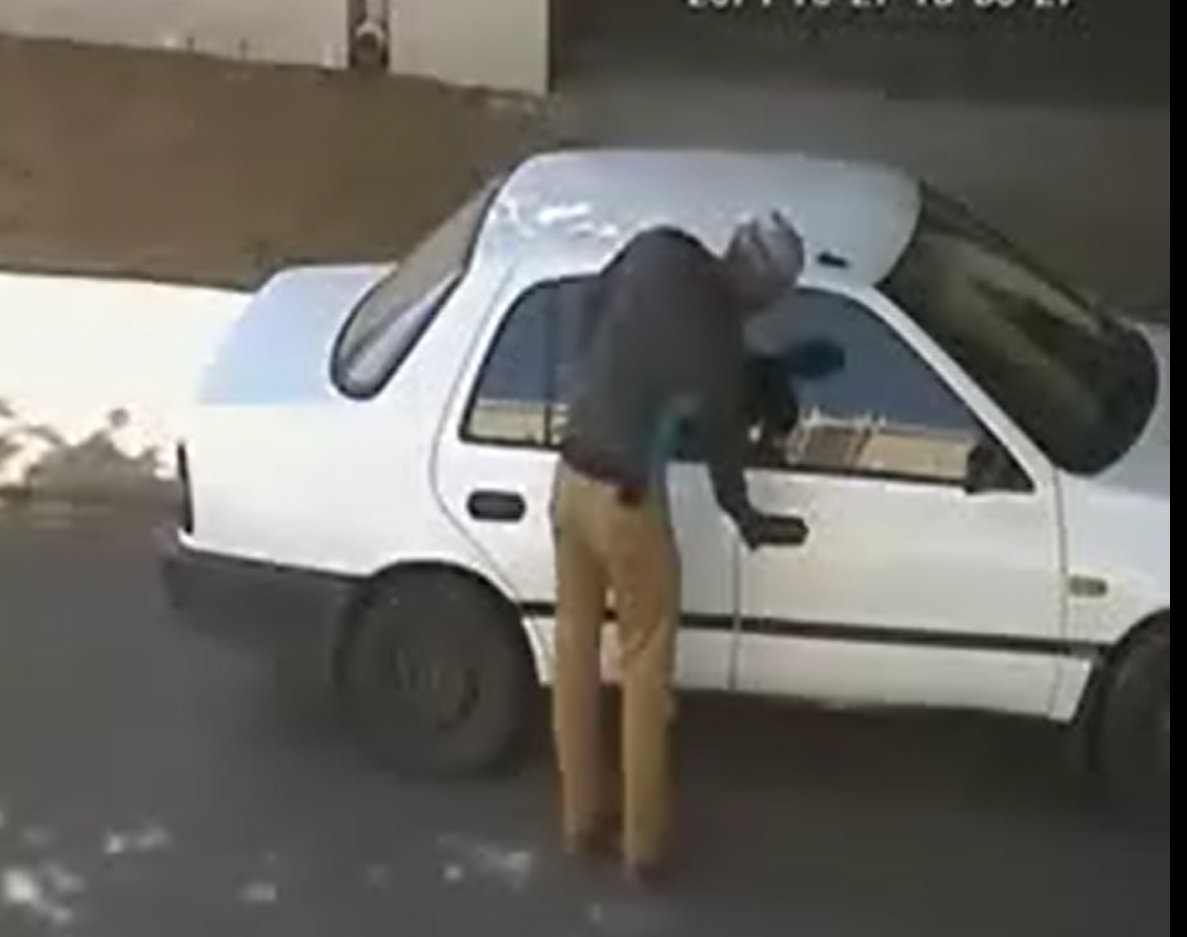}
\end{minipage}
&
\begin{minipage}[t]{\linewidth}\vspace{0pt}
    \setlength{\parskip}{0pt}
    \setlength{\parindent}{0pt}
    \linespread{0.95}\selectfont

    \textbf{VLM Caption:} ``Person standing beside an open driver-side door of a white sedan, holding a long thin object.'' \par
    \textbf{LLM Reasoning:} ``Behavior appears ambiguous; no explicit force or threat observed.'' \par
    \textcolor{red}{\textbf{Prediction: Normal}}
\end{minipage}
&
\begin{minipage}[t]{\linewidth}\vspace{0pt}
    \setlength{\parskip}{0pt}
    \setlength{\parindent}{0pt}
    \linespread{0.95}\selectfont

    \textbf{Agent Query (T1):} \textit{``Is the person deliberately attempting to open or manipulate the door?''} \par
    \textbf{VLM Response:} ``No deliberate manipulation observed.'' \par
    \vspace{1pt} 
    \textbf{Agent Query (T2):} \textit{``Is the object being inserted into the vehicle?''} \par
    \textbf{VLM Response:} ``Yes, deliberate movements into the car are observed.'' \par
    \textbf{LLM Re-eval:} ``Object insertion into vehicle suggests theft or vandalism.'' \par
    \textcolor{blue}{\textbf{Prediction: Anomalous}}
\end{minipage}
\\
\bottomrule
\end{tabular}
\caption{\textbf{Qualitative comparison of Static vs. Dynamic Prompting.} In Turn 0 (standard VLM prompting), the model captures general scene dynamics but misses fine-grained semantic cues, leading to False Negatives. In Turn 1 or on Turn 2, the QVAD Agent hypothesizes a potential anomaly and generates a targeted query, correcting the prediction without parameter updates.}
\label{fig:qualitative_analysis}
\end{figure*}
\begin{figure*}[h]
    \centering
    \includegraphics[width=\textwidth]{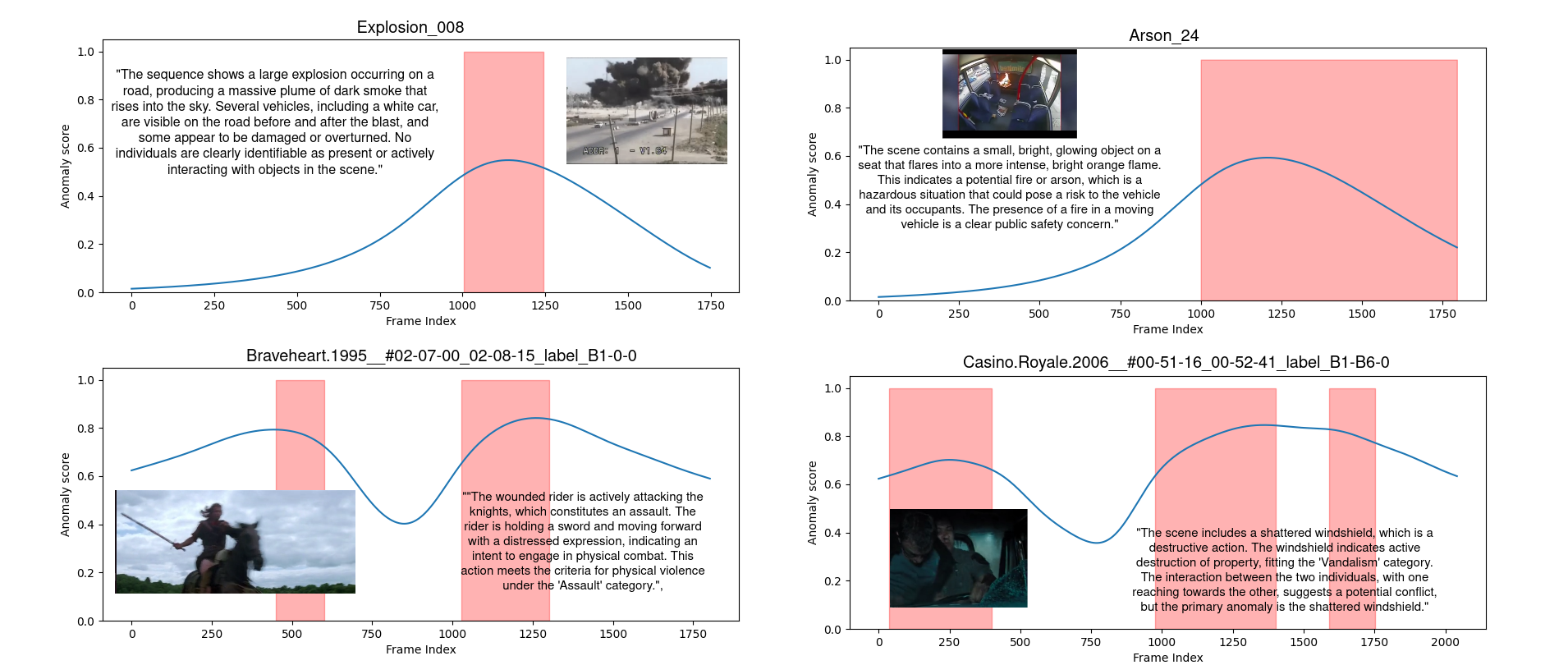}
    \caption{Examples of anomaly scores and reasoning for videos from UCF-Crime and XD-Violence test set.}
    \label{fig: Example}
\end{figure*}
We compare QVAD against state-of-the-art training-free frameworks, including LAVAD, Panda, and VADTree. As shown in Table 1 and Table 2, QVAD achieves competitive or superior performance across all benchmarks while utilizing significantly fewer computational resources.

On UCF-Crime, QVAD achieves an AUC of 84.28\%, effectively matching the performance of Panda and VADTree and outperforming LAVAD (80.28\%) by a significant margin, despite its significantly smaller parameter footprint. Similarly, on XD-Violence, we achieve 68.53\% AP, demonstrating robust detection of violent events.

Our method excels in open-set scenarios. On UBNormal (Table 2), QVAD achieves a state-of-the-art AUC of 79.6\%, surpassing Panda (75.78\%) and LAVAD (64.23\%). This indicates that our dynamic questioning mechanism adapts to "unknown" anomalies better than static prompting approaches, which tend to overfit to standard anomaly definitions. Furthermore, on the challenging ComplexVAD dataset, QVAD achieves 68.02\% AUC, demonstrating its ability to maintain and refine contextual understanding over extended temporal windows. For comparison, we additionally implement VadTree on those datasets, obtaining AUC scores of 65.80\% on UBNormal and 54.20\% on ComplexVAD. Detailed implementation specifics for VadTree are provided in Appendix~\ref{sub: implenVad}.

We further present a qualitative analysis to illustrate the effectiveness of the proposed method. Figure~\ref{fig:qualitative_analysis} demonstrates the efficacy of the agentic framework in QVAD. Although the LLM initially fails to identify the anomaly, the agentic refinement process enables the system to iteratively update its hypotheses and progressively improve its reasoning, ultimately leading to accurate anomaly recognition.

Similarly, Figure~\ref{fig: Example} presents representative examples of anomaly scores \((\hat{p}_t)\) plotted against the frame index, together with the corresponding reasoning \(R_t\), for test videos sampled from the UCF-Crime and XD-Violence datasets. These results reveal a strong correlation between the predicted anomaly score \(\hat{p}_t\) and the generated reasoning \(R_t\), indicating that higher anomaly probability is consistently aligned with more salient, coherent, and semantically meaningful reasoning.

\begin{table}[t]
\caption{Comparison with state-of-the-art methods in terms of model size and estimated GPU memory requirements. LPM denotes the largest parameterized model used by each method.}
\label{tab:sota_comparison}
\centering
\small
\setlength{\tabcolsep}{6pt}
\begin{tabular}{lccc}
\toprule
Method & Total Params (B) & LPM (B) & $\approx$ \text{Min. GPU} \\
\midrule
LAVAD   & 79.60 & 13.0 & 30 (GB) \\
Panda   & N/A   & N/A  & 48 (GB) \\
VAD Tree& 22.20 & 14.0 & 32 (GB) \\
\midrule
\textbf{QVAD (Ours)} & \textbf{8.02} & \textbf{4.0} & \textbf{9 (GB)} \\
\bottomrule
\end{tabular}
\begin{tablenotes}
\small
\item \textit{Note:} The parameter count of the Gemini~2.0~Flash model used in Panda is not publicly available; therefore, we report the GPU memory of the hardware used in their experiments (NVIDIA A6000, 48\,GB). The minimum GPU memory corresponds to a conservative estimate required to safely run the largest model component (LPM) of each method.
\end{tablenotes}
\end{table}


\subsection{Efficiency and Edge Deployment}

The primary contribution of QVAD extends beyond detection accuracy to the efficiency required for practical deployment. In Table~\ref{tab:sota_comparison}, we contrast our approach against state-of-the-art methods that suffer from prohibitive computational costs and latency.

\begin{table}[t]
\caption{Ablation study on UBNormal (UB) AUC with different models and their  GPU memory usage observed with the Memory-Optimized Pipeline.}
\label{tab:ablation3}
\centering
\small
\setlength{\tabcolsep}{5pt}
\begin{tabular}{lcc}
\toprule
Models & QWEN3-VL-4B & QWEN3-VL-2B \\
\midrule
QWEN3-4B    & \textbf{79.60} (8.56 GB)  &  77.76 (8.36 GB) \\ 
QWEN3-1.7B  &  75.92 (8.56 GB) & 73.88 (4.20 GB) \\
\bottomrule
\end{tabular}
\vskip -0.1in
\end{table}
\begin{table}[h]
\caption{Ablation study on UCF-Crime (UCF) AUC with different model combinations under the Memory-Optimized Pipeline.}
\label{tab:ablation_ucf}
\centering
\small
\setlength{\tabcolsep}{5pt}
\begin{tabular}{lcc}
\toprule
Models & QWEN3-VL-4B & QWEN3-VL-2B \\
\midrule
QWEN3-4B    & \textbf{84.28} & 81.51 \\
QWEN3-1.7B  & 83.12 & 83.75 \\
\bottomrule
\end{tabular}
\vskip -0.1in
\end{table}

Our \textbf{Standard Pipeline} operates at an average of 5.98 FPS using the 4B/4B configuration, increasing to 6.19 FPS when utilizing the lighter 1.7B/2B variant on a single NVIDIA RTX 5090 across the entire UCF-Crime dataset. To ensure a fair comparison, it is crucial to contextualize the hardware setups of competing methods:
\begin{itemize}
    \item \textbf{Panda} relies on API-based calls to massive foundation models ($\approx$80B parameters). They report 0.82 FPS which stems from both the immense computational cost of the model and the inevitable network latency. 
    \item \textbf{VADTree}, when evaluated on the same RTX 5090 GPU, we achieved $1.98$ FPS. 
\end{itemize}

We also evaluate the \textbf{Memory-Optimized Pipeline}, specifically designed for hardware with limited VRAM. By sequentially loading and offloading the VLM and LLM, we achieve the minimal GPU memory footprints reported in Table~\ref{tab:sota_comparison} and Table~\ref{tab:ablation3} (e.g., 4.20 GB for the 1.7B/2B variant and $\approx9$ GB for 4B/4B variant). While this swapping process introduces a marginal latency overhead (dropping the average speed to 5.89 FPS for the 4B/4B configuration), the trade-off is negligible. This optimization allows QVAD to run on consumer-grade edge device hardware (e.g., NVIDIA Jetson Orin Nano), 
maintaining real-time capabilities in environments where competitors like VADTree and Panda simply cannot function.

\begin{table}[t]
\caption{Ablation study for number of turns.}
\label{tab:ablation2}
\centering
\small
\setlength{\tabcolsep}{5pt}
\begin{tabular}{lcccc}
\toprule
Turns & UCF  & XD & UB & CVAD \\
       &  (AUC) &  (AP) & (AUC) & (AUC) \\
\midrule
0 &  77.23  &  68.23 &  77.95 &  49.81  \\
1 &  82.18  &  68.73 &  78.20 &  49.67  \\
2 &  84.28  &  68.53 &  79.6 &  68.02  \\
\bottomrule
\end{tabular}
\vskip -0.1in
\end{table}
\begin{table}[ht]
\caption{Ablation study on UCF (AUC)}
\label{tab:ablation_base}
\centering
\small
\setlength{\tabcolsep}{5pt}
\begin{tabular}{lcccc}
\toprule
Configuration & Vec & Sel & UCF (AUC) \\
\midrule
QVAD (Vec+Sel)    & $\checkmark$ & $\checkmark$ & \textbf{84.28} \\
QVAD (Sel)     & $\times$ & $\checkmark$  & 82.35 \\
QVAD (Base)       & $\times$ & $\times$ &  81.67 \\
\bottomrule
\end{tabular}
\vskip -0.2in
\end{table}
\subsection{Ablation Study}

\textbf{Efficiency vs. Performance.} To further demonstrate the viability of QVAD for resource-constrained edge devices, we evaluate the performance of smaller backbone configurations. Table~\ref{tab:ablation3} and Table \ref{tab:ablation_ucf} present the results of deploying QVAD with compact model variants (Qwen3-1.7B and Qwen3-VL-2B).

Remarkably, the 1.7B/2B configuration retains strong performance (83.75\% and 73.88\% AUC on UCF Crime and UBNormal respectively) while requiring only 4.20 GB of GPU memory using our Memory-Optimized pipeline. This footprint is well within the specifications of embedded AI computers such as the NVIDIA Jetson Orin Nano (8GB), validating QVAD as a feasible solution for real-time, on-device video anomaly detection.
Although the performance of 1.7B/2B configuration seems to be unexpectedly a little higher than the 1.7B/4B and 4B/2B versions, the difference is statistically insignificant. 

\textbf{LLM-VLM Dialogue Turns.} The ablation study on dialogue iterations (Table~\ref{tab:ablation2}) substantiates the core hypothesis of our agentic framework: the VLM's detection capability is strictly enhanced by the LLM's guided inquiries. We observe a clear performance progression from the static baseline (Turn 0) to the refined state (Turn 2), though the magnitude of improvement is contingent on the anomaly's complexity.

For XD-Violence, the gains are marginal, suggesting a ``saturation effect''. The anomalies in this domain (e.g., explosions, fighting) are visually salient, allowing the VLM to resolve them with minimal reasoning depth. In stark contrast, in ComplexVAD a non-linear performance jump is observed. Accuracy remains stagnant at Turn 1 but surges dramatically at Turn 2 (from $\approx$50\% to 68\% AUC). This indicates that for subtle, long-term anomalies, a single exploratory question is insufficient; the agent requires a multi-step deductive chain to formulate a precise hypothesis that isolates the irregularity.

We limit the interaction to Turn 2 as a strategic design choice. Extending the dialogue further induces substantial growth in the Key-Value (KV) cache, resulting in prohibitive latency and memory overhead.

\textbf{Impact of Vector Memory and Selection.} We first isolate the contributions of our core components: the Vector Memory module (Vec), which retains long-term temporal context, and the motion aware frame selection (Sel), which filters irrelevant frames or queries. As shown in the first section of Table 6, the baseline model achieves 81.67\% AUC when tested on UCF-Crime. Introducing the Selection Mechanism improves performance to 82.35\%, validating that filtering noise is beneficial. However, the integration of the Vector Memory yields the most significant gain, boosting AUC to 84.28\%. This confirms that for long-form video anomaly detection, immediate visual context is insufficient; the ability to recall and correlate past events via vector-based retrieval is essential for identifying complex anomalies.

\section{Discussion}

\textbf{Potential for Acceleration via KV Caching.}
While our current implementation utilizes standard autoregressive inference, the iterative structure of our framework is uniquely positioned to capitalize on emerging research in Key-Value (KV) cache optimization. A substantial computational cost in multi-turn VLM systems is the redundant re-processing of high-dimensional visual features ($\mathbf{X}_t$) at every dialogue turn. Our method is architecturally aligned with recent advances in efficient attention mechanisms, such as PagedAttention \cite{kwon2023efficient}. By persisting the attention states of the visual encoding and dialogue history $\mathcal{H}_t$, future implementations could eliminate the quadratic cost of re-tokenizing the context window. This implies that as the domain of memory-efficient inference matures, our agentic approach will naturally scale to achieve significantly lower latency compared to static prompting baselines, where such state persistence is less effective.

\textbf{Scalable Context with Vector Spaces.}
Furthermore, the integration of a vector-based memory module aligns with the growing trend of retrieval-augmented generation (RAG) in efficient ML. By offloading long-term temporal dependencies to a compressed vector space rather than extending the active context window, we maintain a bounded memory footprint even as video duration increases. This combination of retrieval-based long-term history and the potential for cached short-term dialogue offers a promising path toward real-time, long-context video understanding on resource-constrained hardware.

\section{Conclusion}

We presented QVAD, a question-centric agentic framework for training-free video anomaly detection that addresses the limitations of static prompting with smaller models through dynamic, context-aware query refinement. Our approach achieves state-of-the-art performance on multiple benchmarks through iterative question-answer interactions between LLM and VLM agents, while remaining practical for deployment on edge devices due to its significantly lower parameter count.

\newpage

\section*{Impact Statement}

This paper presents work whose goal is to advance the field of Machine Learning, specifically in video anomaly detection. Our training-free approach reduces computational requirements and improves accessibility to advanced VAD capabilities. While our method has potential applications in surveillance and security, we acknowledge the importance of responsible deployment with appropriate privacy safeguards and human oversight. There are many potential societal consequences of our work, none which we feel must be specifically highlighted here beyond the standard considerations for video analysis systems.


\bibliography{example_paper}

@inproceedings{zanella2024lavad,
  author    = {L. Zanella and W. Menapace and M. Mancini and Y. Wang and E. Ricci},
  title     = {Harnessing Large Language Models for Training-Free Video Anomaly Detection},
  booktitle = {Proceedings of the IEEE/CVF Conference on Computer Vision and Pattern Recognition (CVPR)},
  pages     = {18527--18536},
  year      = {2024}
}

@article{yang2025panda,
  author    = {Z. Yang and C. Gao and M. Z. Shou},
  title     = {{PANDA}: Towards Generalist Video Anomaly Detection via Agentic {AI} Engineer},
  journal   = {arXiv preprint arXiv:2509.26386},
  year      = {2025}
}

@article{li2025vadtree,
  author    = {W. Li and Y. Xu and Y. Rao and Z. Wang and S. Deng},
  title     = {{VADTree}: Explainable Training-Free Video Anomaly Detection via Hierarchical Granularity-Aware Tree},
  journal   = {arXiv preprint arXiv:2510.22693},
  year      = {2025}
}

@inproceedings{ye2025vera,
  author    = {M. Ye and W. Liu and P. He},
  title     = {{VERA}: Explainable Video Anomaly Detection via Verbalized Learning of Vision-Language Models},
  booktitle = {Proceedings of the IEEE/CVF Conference on Computer Vision and Pattern Recognition (CVPR)},
  pages     = {8679--8688},
  year      = {2025}
}

@inproceedings{wu2024ovvad,
  author    = {P. Wu and X. Zhou and G. Pang and Y. Sun and J. Liu and P. Wang and Y. Zhang},
  title     = {Open-Vocabulary Video Anomaly Detection},
  booktitle = {Proceedings of the IEEE/CVF Conference on Computer Vision and Pattern Recognition (CVPR)},
  pages     = {18297--18307},
  year      = {2024}
}

@inproceedings{yang2024rules,
  author    = {Y. Yang and K. Lee and B. Dariush and Y. Cao and S. Y. Lo},
  title     = {Follow the Rules: Reasoning for Video Anomaly Detection with Large Language Models},
  booktitle = {European Conference on Computer Vision (ECCV)},
  pages     = {304--322},
  year      = {2024},
  publisher = {Springer Nature Switzerland},
  address   = {Cham}
}

@inproceedings{wu2024vadclip,
  author    = {P. Wu and X. Zhou and G. Pang and L. Zhou and Q. Yan and P. Wang and Y. Zhang},
  title     = {{VADCLIP}: Adapting Vision-Language Models for Weakly Supervised Video Anomaly Detection},
  booktitle = {Proceedings of the AAAI Conference on Artificial Intelligence},
  volume    = {38},
  number    = {6},
  pages     = {6074--6082},
  year      = {2024}
}

@inproceedings{zhang2025holmes,
  author    = {H. Zhang and X. Xu and X. Wang and J. Zuo and X. Huang and C. Gao and N. Sang},
  title     = {{Holmes-VAU}: Towards Long-Term Video Anomaly Understanding at Any Granularity},
  booktitle = {Proceedings of the IEEE/CVF Conference on Computer Vision and Pattern Recognition (CVPR)},
  pages     = {13843--13853},
  year      = {2025}
}

@inproceedings{wu2020multimodal,
  author    = {P. Wu and J. Liu and Y. Shi and Y. Sun and F. Shao and Z. Wu and Z. Yang},
  title     = {Not Only Look, But Also Listen: Learning Multimodal Violence Detection Under Weak Supervision},
  booktitle = {European Conference on Computer Vision (ECCV)},
  pages     = {322--339},
  year      = {2020}
}

@inproceedings{tian2021rtfm,
  author    = {Y. Tian and G. Pang and Y. Chen and R. Singh and J. W. Verjans and G. Carneiro},
  title     = {Weakly-Supervised Video Anomaly Detection with Robust Temporal Feature Magnitude Learning},
  booktitle = {Proceedings of the IEEE/CVF International Conference on Computer Vision (ICCV)},
  pages     = {4975--4986},
  year      = {2021}
}

@article{hemmatyar2025hycovad,
  author  = {M. M. Hemmatyar and M. Jafari and M. A. Yousefi and M. R. Nemati and M. Azadani and H. R. Rastad and A. Akbari},
  title   = {HyCoVAD: A Hybrid SSL-LLM Model for Complex Video Anomaly Detection},
  journal = {arXiv preprint arXiv:2509.22544},
  year    = {2025}
}

@inproceedings{mumcu2025complexvad,
  author    = {F. Mumcu and M. Jones and Y. Yilmaz and A. Cherian},
  title     = {ComplexVAD: Detecting Interaction Anomalies in Video},
  booktitle = {Proceedings of the Winter Conference on Applications of Computer Vision (WACV)},
  pages     = {1093--1102},
  year      = {2025}
}

@inproceedings{sultani2018real,
  author    = {W. Sultani and C. Chen and M. Shah},
  title     = {Real-World Anomaly Detection in Surveillance Videos},
  booktitle = {Proceedings of the IEEE Conference on Computer Vision and Pattern Recognition (CVPR)},
  pages     = {6479--6488},
  year      = {2018}
}

@inproceedings{wu2020not,
  author    = {P. Wu and J. Liu and Y. Shi and Y. Sun and F. Shao and Z. Wu and Z. Yang},
  title     = {Not Only Look, But Also Listen: Learning Multimodal Violence Detection Under Weak Supervision},
  booktitle = {European Conference on Computer Vision (ECCV)},
  pages     = {322--339},
  year      = {2020}
}

@inproceedings{acsintoae2022ubnormal,
  author    = {A. Acsintoae and A. Florescu and M. I. Georgescu and T. Mare and P. Sumedrea and R. T. Ionescu and F. S. Khan and M. Shah},
  title     = {{Ubnormal}: New Benchmark for Supervised Open-Set Video Anomaly Detection},
  booktitle = {Proceedings of the IEEE/CVF Conference on Computer Vision and Pattern Recognition (CVPR)},
  pages     = {20143--20153},
  year      = {2022}
}

@article{qi2025time,
  title={Time-efficient Video Anomaly Detection with Parallel Computing and Twice-reconstruction},
  author={Qi, Xiaosha and Chao, Xin and Ji, Genlin and Li, Le},
  journal={IEEE Sensors Journal},
  year={2025},
  publisher={IEEE}
}

@inproceedings{zhang2024multi,
  title={Multi-scale video anomaly detection by multi-grained spatio-temporal representation learning},
  author={Zhang, Menghao and Wang, Jingyu and Qi, Qi and Sun, Haifeng and Zhuang, Zirui and Ren, Pengfei and Ma, Ruilong and Liao, Jianxin},
  booktitle={Proceedings of the IEEE/CVF Conference on Computer Vision and Pattern Recognition},
  pages={17385--17394},
  year={2024}
}

@inproceedings{al2024collaborative,
  title={Collaborative learning of anomalies with privacy (clap) for unsupervised video anomaly detection: A new baseline},
  author={Al-Lahham, Anas and Zaheer, Muhammad Zaigham and Tastan, Nurbek and Nandakumar, Karthik},
  booktitle={Proceedings of the IEEE/CVF Conference on Computer Vision and Pattern Recognition},
  pages={12416--12425},
  year={2024}
}

@inproceedings{ristea2024self,
  title={Self-distilled masked auto-encoders are efficient video anomaly detectors},
  author={Ristea, Nicolae-C and Croitoru, Florinel-Alin and Ionescu, Radu Tudor and Popescu, Marius and Khan, Fahad Shahbaz and Shah, Mubarak and others},
  booktitle={Proceedings of the IEEE/CVF conference on computer vision and pattern recognition},
  pages={15984--15995},
  year={2024}
}

@inproceedings{karim2024real,
  title={Real-time weakly supervised video anomaly detection},
  author={Karim, Hamza and Doshi, Keval and Yilmaz, Yasin},
  booktitle={Proceedings of the IEEE/CVF winter conference on applications of computer vision},
  pages={6848--6856},
  year={2024}
}

@inproceedings{chen2024prompt,
  title={Prompt-enhanced multiple instance learning for weakly supervised video anomaly detection},
  author={Chen, Junxi and Li, Liang and Su, Li and Zha, Zheng-jun and Huang, Qingming},
  booktitle={Proceedings of the IEEE/CVF Conference on Computer Vision and Pattern Recognition},
  pages={18319--18329},
  year={2024}
}

@article{mumcu2025leveraging,
  title={Leveraging Multimodal LLM Descriptions of Activity for Explainable Semi-Supervised Video Anomaly Detection},
  author={Mumcu, Furkan and Jones, Michael J and Cherian, Anoop and Yilmaz, Yasin},
  journal={arXiv preprint arXiv:2510.14896},
  year={2025}
}

@inproceedings{shao2025eventvad,
  author    = {Y. Shao and H. He and S. Li and S. Chen and X. Long and F. Zeng and S. Li},
  title     = {EventVAD: Training-Free Event-Aware Video Anomaly Detection},
  booktitle = {Proceedings of the 33rd ACM International Conference on Multimedia},
  pages     = {2586--2595},
  year      = {2025}
}

@article{bai2025qwen3vl,
  author  = {S. Bai and Y. Cai and R. Chen and K. Chen and X. Chen and Z. Cheng and K. Q. Zhu},
  title   = {Qwen3-VL Technical Report},
  journal = {arXiv preprint arXiv:2511.21631},
  year    = {2025}
}

@article{yang2025qwen3,
  author  = {A. Yang and A. Li and B. Yang and B. Zhang and B. Hui and B. Zheng and Z. Qiu},
  title   = {Qwen3 Technical Report},
  journal = {arXiv preprint arXiv:2505.09388},
  year    = {2025}
}

@inproceedings{kwon2023efficient,
  title={Efficient memory management for large language model serving with pagedattention},
  author={Kwon, Woosuk and Li, Zhuohan and Zhuang, Siyuan and Sheng, Ying and Zheng, Lianmin and Yu, Cody Hao and Gonzalez, Joseph and Zhang, Hao and Stoica, Ion},
  booktitle={Proceedings of the 29th symposium on operating systems principles},
  pages={611--626},
  year={2023}
}
\bibliographystyle{icml2026}

\newpage
\appendix
\onecolumn

\begin{table}[h]
\caption{Ablation study on window size and stride on UCF (AUC)}
\label{tab:ablation_window}
\vskip 0.01in
\centering
\small
\setlength{\tabcolsep}{5pt}
\begin{tabular}{lcc}
\toprule
Configuration & Window / Stride & UCF (AUC) \\
\midrule
QVAD & 64 / 32   & 81.92 \\
QVAD & 128 / 64  & \textbf{84.28} \\
QVAD & 256 / 128 & 83.00 \\
\bottomrule
\end{tabular}
\vskip -0.1in
\end{table}
\begin{table}[h]
\caption{Ablation study on number of frames on UCF (AUC)}
\label{tab:ablation_frames}
\vskip 0.01in
\centering
\small
\setlength{\tabcolsep}{5pt}
\begin{tabular}{lcc}
\toprule
Configuration & \# Frames & UCF (AUC) \\
\midrule
QVAD & 4  & 79.30 \\
QVAD & 8  & \textbf{84.28} \\
QVAD & 16 & 83.10 \\
\bottomrule
\end{tabular}
\vskip -0.1in
\end{table}
\begin{table}[h]
\caption{Ablation study on confidence threshold $\theta$ (AUC UBnormal)}
\label{tab:ablation_theta}
\vskip 0.01in
\centering
\small
\setlength{\tabcolsep}{10pt}
\begin{tabular}{lc}
\toprule
Confidence Threshold $\theta$ & AUC UBnormal (\%) \\
\midrule
0.6 & 66.30 \\
0.7 & \textbf{79.60} \\
0.8 & 65.56 \\
\bottomrule
\end{tabular}
\end{table}

\section{Additional Experimental Results}

To provide a comprehensive understanding of the QVAD framework, we conduct extended ablation studies focusing on component contributions, hyperparameter sensitivity, and the stochastic stability of the agentic dialogue.
\subsection{Component Analysis and Hyperparameter Sensitivity}
\textbf{Temporal Granularity (Window Size).} We analyze the impact of the temporal window size on detection performance. A window size of \textbf{128 frames} (stride 64) proves optimal (84.28\% AUC). Smaller windows (64 frames) fail to capture sufficient temporal context for slowly evolving anomalies (81.92\%), while larger windows (256 frames) introduce excessive irrelevant information, diluting the anomaly signal (83.00\%).

\textbf{Visual Density (Number of Frames).} Similarly, we examine the number of frames sampled per window. We find that \textbf{8 frames} provide the optimal balance between information density and reasoning clarity. Using only 4 frames results in a significant performance drop (79.30\%), as the VLM misses critical motion cues. Conversely, increasing the sample to 16 frames (83.10\%) degrades performance, likely due to information redundancy overwhelming the small VLM's context window without adding semantic value.

\textbf{Decision Boundary (Confidence Threshold $\theta$).} Finally, we calibrate the confidence threshold $\theta$ that governs the agent's uncertainty mechanism. We select UBnormal for this ablation due to its open-set nature and diverse, subtle anomaly types, which demand a more strictly tuned decision boundary than the often unambiguous crimes in UCF. We observe a distinct performance peak at $\theta$ = 0.7 (79.60\% AUC). Deviating from this optimum leads to sharp declines: a lower threshold ($\theta=0.6$) likely introduces noise by over-querying benign events, while a stricter threshold ($\theta=0.8$) suppresses necessary clarifications for ambiguous anomalies.
\newpage
\subsection{Stability and Reproducibility}
A common concern with LLM-based agents is the stochastic nature of text generation, which can lead to variance in reasoning paths. To evaluate the robustness of QVAD, we conducted a stability analysis across three independent runs (T1--T3) on both the UCF Crime and UBnormal datasets. As detailed in Table~\ref{tab:ablation_stability}, QVAD demonstrates remarkable stability across both benchmarks. On UCF Crime, which features more clear
anomalies, the system yields a mean AUC of 84.03\% with a minimal standard deviation of 0.204. Crucially, this stability holds even on the UBnormal dataset, whose subtle, open-set anomalies make it a rigorous stress test for generation sensitivity. Here, QVAD achieves a mean AUC of 79.20\% with a standard deviation of only 0.283. This consistently low variance ($\sigma < 0.3$) confirms that our ``Guiding Question'' refinement process converges to reliable reasoning paths, effectively mitigating the inherent non-determinism of the underlying LLM.
\begin{table}[h]
\caption{Stability Analysis on UBnormal (AUC) and UCF (AUC)}
\label{tab:ablation_stability}
\vskip 0.01in
\centering
\small
\setlength{\tabcolsep}{5pt}
\begin{tabular}{lccccc}
\toprule
Dataset & T1 & T2 & T3 & Mean & STD \\
\midrule
UCF & 83.78 & \textbf{84.28} & 84.02 & 84.03 & 0.204\\
UB & 78.98  &  \textbf{79.60} &  79.02  & 79.20 & 0.283 \\
\bottomrule
\end{tabular}
\vskip -0.5in
\end{table}
\section{Implementation Details}
\label{sec:implementation}
This section provides comprehensive algorithmic details of our video anomaly detection pipeline, including the overall architecture, vector memory system, motion-aware frame selection, and the prompting strategies for both the Vision-Language Model (VLM) and Large Language Model (LLM).

\subsection{Overall Pipeline}
\label{subsec:pipeline}

Our system employs a multi-turn conversational framework between a VLM and an LLM, augmented with semantic vector memory for temporal context retrieval. Algorithm~\ref{alg:main_pipeline} describes the complete anomaly detection workflow.

\noindent The \textsc{Enrich} function concatenates the initial caption, Q\&A pairs, and retrieved context under a strict token budget $\tau_{\max} = 2048$.
\subsection{Motion-Aware Frame Selection}
\label{subsec:motion}

To efficiently process video while preserving anomaly-relevant content, we employ a two-stage frame selection strategy. First, we uniformly sample 32 frames from the input window to reduce computational load. Then, we apply motion-aware selection to identify the 8 most informative frames, always including the first and last frames for temporal anchoring.

\subsection{Vector Memory System}
\label{subsec:memory}

To enable temporal reasoning across non-adjacent video segments, we implement a semantic vector memory using sentence embeddings. Each processed scene is stored with its embedding, enabling efficient retrieval of contextually relevant historical observations.

\noindent\textbf{Parameters:} Max memory size $M_{\max} = 400$, max context tokens $\tau_{\text{ctx}} = 512$, max caption tokens $\tau_{\text{cap}} = 150$, retrieval count $k = 3$. We use MiniLM-L6-v2 for the encoder $\mathbf{E}$. This choice is critical for our edge-deployment objective: the model provides high-fidelity semantic matching while occupying less than 100MB of memory and offering significantly higher throughput than standard BERT-based encoders. This ensures the retrieval step imposes negligible latency overhead ($\approx$ 2ms per query) on the overall inference pipeline.

\begin{algorithm}[H]
\caption{Video Anomaly Detection Pipeline}
\label{alg:main_pipeline}
\begin{algorithmic}[1]
\Require Video windows $\mathcal{W}_t$, maximum turns $K$
\Ensure Anomaly results $\{p_t, g_t, R_t\}$
\State \textbf{Initialize:} $\mathcal{M}_v \gets \text{VectorMemory}()$
\For{$\mathcal{W}_t \in \mathcal{V}$}
    \State $\mathcal{S}_{uni} \gets \textsc{UniformSample}(\mathcal{W}_t, 32)$
    \State $\mathbf{X_t} \gets \textsc{MotionSelect}(\mathcal{S}_{uni}, 8)$ \Comment{Alg.~\ref{alg:motion}}
    \Statex \textbf{// Initial captioning and scoring}
    \State $\mathcal{H}_t \gets [\,]$, \quad $t \gets 0$, \quad $k \gets 0$
    \State $C_t \gets \mathcal{P}(\mathbf{X_t}, P_{\text{init}})$
    \State $({g}_t^k, {p}_t^k, {R}_t^k) \gets \mathcal{A}(C_t, \mathcal{H}_t)$
    \Statex \textbf{// Iterative refinement loop}
    \While{$k < K$ \textbf{and} $p_t < 0.7$}
        \State ${q}_t^{k+1} \gets \mathcal{A} (C_t, \mathcal{H}_t, {R}_t^k, {g}_t^k, {p}_t^k)$
        \State $\text{ctx} \gets \textsc{RetrieveContext}(q_t^{k+1}, \mathcal{M}_v)$ \Comment{Alg.~\ref{alg:memory}}
        \State ${a}_t^{k+1} \gets \mathcal{P}(\mathbf{X_t},{q}_t^{k+1}, P_{\text{focus}})$
        \State $\mathcal{H}_t.\text{append}((q_t^{k+1}, a_t^{k+1}))$
        \State $(g_t^{k+1},p_t^{k+1},R_t^{k+1}) \gets \mathcal{A}(C_t,\mathcal{H}_t,\text{ctx})$
        \State $k \gets k + 1$
    \EndWhile
    \Statex \textbf{// Update memory and collect results}
    \State $\mathcal{M}_v.\text{add}(C_t,g_t^k,\mathcal{H}_t,\text{ctx})$  
\EndFor
\State \Return $\{p_t^k,g_t^k,R_t^k\}$
\end{algorithmic}
\end{algorithm}

\begin{algorithm}[H]
\caption{Vector Memory Operations}
\label{alg:memory}
\begin{algorithmic}[1]

\Ensure: Encoder $\mathbf{E}$, index $\mathcal{I}$, metadata store $\mathbf{M_v}$
\Procedure{AddScene}{$C_t$, $g^k_t$, $\mathcal{H}_t$, $ctx$}
    \State $c' \gets \textsc{Truncate}(\text{($C_t$ + $\mathcal{H}_t$ + $ctx$)}, \tau_{\text{cap}})$
    \State $\mathbf{e} \gets \mathbf{E}(c')$\Comment{$\mathbf{e} \in \mathbb{R}^{384}$}
    \State $\mathcal{I}.\text{add}(\mathbf{e})$
    \State $\mathbf{M_v}.\text{append}( c', g_t)$
    \If{$|\mathbf{M_v}| > M_{\max}$}
        \State \textsc{RebuildIndex}() \Comment{Prune to /2}
    \EndIf
\EndProcedure
\Procedure{RetrieveContext}{query, $k$}
    \If{$|\mathcal{I}| = 0$}
        \Return \texttt{None}
    \EndIf
    \State $\mathbf{q}_e \gets \mathbf{E}(\text{query})$
    \State $\mathbf{J} \gets \mathcal{I}.\text{search}(\mathbf{q}_e, top-$K$)$ \Comment{top-$K$}
    \State $\text{result} \gets [\,]$, \quad $\text{tokens} \gets 0$
    \For{$j \in \mathbf{J}$}
        \State $\text{entry}  \gets \mathbf{M_v}(j)$
        \State $\text{result}.\text{append}(\text{entry})$
        \State $\text{tokens} \gets \text{tokens} + |\text{entry}|$
        \If{$\text{tokens} > \tau_{\text{ctx}}$}
            \textbf{break}
        \EndIf
    \EndFor
    \State \Return $\textsc{Join}(\text{result})$
\EndProcedure
\end{algorithmic}
\end{algorithm}

\begin{algorithm}[H]
\caption{Motion-Aware Frame Selection}
\label{alg:motion}
\begin{algorithmic}[1]
\Require Frames $\mathcal{S}_{uni} = \{f_1, \ldots, f_N\}$, target count $T$
\For{$i = 1$ \textbf{to} $N-1$}
    \State $g_i \gets \textsc{Grayscale}(f_i)$
    \State $g_{i+1} \gets \textsc{Grayscale}(f_{i+1})$
    \State $\tilde{g}_i \gets \textsc{GaussianBlur}(g_i, \sigma)$
    \State $\tilde{g}_{i+1} \gets \textsc{GaussianBlur}(g_{i+1}, \sigma)$
    \State $D \gets |\tilde{g}_{i+1} - \tilde{g}_i|$
    \State $M \gets \mathbb{I}[D > \delta]$ \Comment{Binary motion mask}
    \State $m_i \gets \text{mean}(M)$ \Comment{Normalized motion score}
\EndFor
\State $m_N \gets m_{N-1}$ \Comment{Copy last score}
\Statex \hrulefill
\State \textbf{Select frames:}
\State $\mathcal{I} \gets \{1, N\}$ \Comment{Always include first/last}
\State $\mathbf{idx} \gets \textsc{ArgSort}(\{m_i\}, \text{descending})$
\For{$j \in \mathbf{idx}$}
    \If{$|\mathcal{I}| \geq T$} \textbf{break}
    \EndIf
    \State $\mathcal{I} \gets \mathcal{I} \cup \{j\}$
\EndFor
\State \Return $\{f_i : i \in \textsc{Sort}(\mathcal{I})\}$
\end{algorithmic}
\end{algorithm}

\subsection{VLM Prompting Strategy}
\label{subsec:vlm_prompts}

We employ two distinct prompting strategies for the VLM: an initial comprehensive prompt and a focused follow-up prompt.

\subsubsection{Initial Caption Prompt}
The initial prompt elicits a structured, objective description of the surveillance footage:

\begin{figure}[h]
\centering
\fbox{\parbox{0.92\columnwidth}{\small
\textbf{VLM Initial Prompt:}\\[0.5em]
Analyze this surveillance video sequence. Provide a concise and objective description. Provide a very brief, 2-3 sentence summary of the main event.\\[0.3em]
1. \textbf{Persons:} How many people are present?\\
2. \textbf{Key Actions:} Describe the specific, atomic actions of each person (e.g., ``walking'', ``standing'', ``fighting'', ``lifting box'') or objects like cars.\\
3. \textbf{Object Interactions:} What objects is the person (or people) interacting with, and how?\\
4. \textbf{Key Visual Details:} Note prominent clothing, visible text, or significant nearby objects (like vehicles).\\[0.3em]
Focus ONLY on what you clearly see. Do not interpret intent or state what is not happening.
}}
\end{figure}

\subsubsection{Focused Response Prompt}
For follow-up questions, we use a constrained prompt to elicit brief, targeted responses:

\begin{figure}[h]
\centering
\fbox{\parbox{0.92\columnwidth}{\small
\textbf{VLM Focused Prompt:}\\[0.5em]
Answer this specific question concisely based on the video frames:\\[0.3em]
\textit{\{question\}}\\[0.3em]
1. Answer directly in 1 single sentence if possible (max 2).\\
2. Be extremely brief.\\
3. Do not describe the whole scene, only answer the specific question.
}}
\end{figure}

\subsection{LLM Prompting Strategy}
\label{subsec:llm_prompts}
The LLM is employed in two complementary roles: (i) anomaly scoring and (ii) clarifying question generation. While the prompt used for question generation remains fixed across datasets, the anomaly scoring prompt must be adapted to each benchmark due to dataset-specific definitions of anomalous events. In particular, the scope of what constitutes an anomaly and the criteria for assigning an anomaly label (Anomaly = 1) are explicitly customized for each dataset.

\subsubsection{Anomaly Scoring System Prompt}
\begin{figure}[h]
\centering
\fbox{\parbox{0.92\columnwidth}{\small
\textbf{UCF Crime LLM Anomaly Scoring System Prompt:}\\[0.5em]
You are a highly analytical OBJECTIVE Public Safety \& Anomaly Detection AI. Your function is to assess a surveillance video caption for the presence of CRIMES or HAZARDOUS ANOMALIES.\\[0.3em]
\textbf{CRITICAL:} BE CONSERVATIVE -- Normal activity is FAR MORE COMMON than anomalies.\\[0.3em]
\textbf{STRICT RULES FOR ANOMALY SCORING:}\\
1. \textbf{SCOPE:} Focus ONLY on: Abuse, Arrest, Arson, Assault, RoadAccidents, Shooting, Shoplifting, Stealing, Vandalism, Fighting, Explosion.\\
2. \textbf{FACT-BASED:} Analyze the EXPLICIT VISUAL FACTS in the caption. Do not hallucinate danger.\\
3. \textbf{ANOMALY = 1 CRITERIA:}\\
\quad A) HUMAN CRIMINALITY: Punches, kicks, weapons drawn, concealing items, deliberate property damage.\\
\quad B) HAZARDOUS SITUATIONS: Vehicle collisions, abnormal positioning, visible wreckage, explosions, thick smoke, open flames.\\
4. \textbf{No Anomaly = 0:} Normal traffic, pedestrians walking/running, light fog, regular parking.\\
5. \textbf{CONFIDENCE:} High (0.8-1.0): explicit crime keywords; Medium (0.5-0.79): suggestive descriptions; Low (0.0-0.49): mostly normal.\\[0.3em]
Respond ONLY in JSON format:\\
\texttt{\{``anomaly\_score'': 0|1, ``confidence'': 0.0-1.0, ``reasoning'': ``...'', ``crime\_type'': ``...''\}}
}}
\end{figure}

\subsubsection{Question Generation Prompt}

\begin{figure}[h]
\centering
\fbox{\parbox{0.92\columnwidth}{\small
\textbf{LLM Question Generation Prompt:}\\[0.5em]
You are an expert surveillance analyst. When given a video description with uncertain or ambiguous elements, you generate ONE specific, focused question to clarify whether criminal activity is occurring.\\[0.3em]
Your questions should:\\
1. Target specific ambiguous details (objects, actions, interactions)\\
2. Help distinguish between normal and criminal behavior\\
3. Be answerable by observing the video footage\\
4. Focus on concrete, observable elements\\
5. Do not ask questions that force VLM to hallucinate a crime\\[0.3em]
If the question needs historical context (e.g., ``Are there patterns?''), phrase it naturally -- the system will automatically provide relevant context.\\[0.3em]
Generate ONLY the question text, no additional commentary.
}}
\end{figure}

\subsubsection{Anomaly Definition on Other Datasets}
To accommodate varying definitions of ``anomaly,'' we utilize dataset-specific instructions.

\begin{figure}[h]
\centering
\setlength{\fboxsep}{3pt}
\fbox{\parbox{0.98\columnwidth}{\scriptsize
\textbf{ComplexVAD System Prompt Definitions:}\\[0.2em]
\textbf{CRITICAL:} Judge ONLY explicit actions. Ignore intent/legality/realness. Anomaly = \textbf{deviation from normal object-object or human-object interaction}.\\[0.2em]
\textbf{Scoring Rules:}
\begin{itemize} \setlength{\itemsep}{0pt} \setlength{\parskip}{0pt} \setlength{\parsep}{0pt}
    \item \textbf{Anomaly = 1:} Clearly abnormal interaction, collision, abandonment, blocking, autonomous object motion, or unusual path/stop.
    \item \textbf{Anomaly = 0:} Normal movement, normal object use, or ambiguous activity.
\end{itemize}
\textbf{Examples of Anomaly = 1:} 1. Person leaves object and walks away. 2. Unexpected collision. 3. Person blocks vehicle. 4. Object moves without human. 5. Sudden unusual stop/turn. 6. Unattended objects near road. 7. Vandalism, theft, assault, arson.\\[0.2em]
\textbf{Examples of Anomaly = 0:} 1. People walking/running/standing normally. 2. Normal use of objects.
}}
\vspace{-0.5em}
\caption{Exact anomaly scoring criteria for ComplexVAD.}
\label{fig:complexvad_prompt}
\end{figure}

\begin{figure*}[h]
\centering
\setlength{\fboxsep}{3pt}
\fbox{\parbox{0.98\textwidth}{\scriptsize
\textbf{XD Violence System Prompt Definitions:}\\[0.2em]
\textbf{CRITICAL PRINCIPLES:} 1. \textbf{CONTEXT-AGNOSTIC:} Source is IRRELEVANT. Score ONLY action. 2. \textbf{MULTIMODAL:} Consider visual AND audio cues. 3. \textbf{TEMPORAL:} ACTIVE violent events only.\\[0.2em]
\textbf{XD-VIOLENCE TAXONOMY - ANOMALY = 1 CRITERIA:}
\begin{itemize} \setlength{\itemsep}{0pt} \setlength{\parskip}{0pt} \setlength{\parsep}{0pt}
    \item \textbf{A) FIGHTING:} Hand-to-hand (punch/kick/grapple); Group brawls; Martial arts (causing harm); Physical assault; Audio: Impact sounds/grunting.
    \item \textbf{B) SHOOTING:} Firearm pointed with intent; Muzzle flash/discharge; Person shot/falling; Active shooter; Audio: Gunshots.
    \item \textbf{C) EXPLOSION:} Visible detonation/blast wave; Large fire/smoke plumes; Debris clouds; Vehicle/building explosions; Audio: Loud bangs.
    \item \textbf{D) RIOT:} Crowd violence/mob attacks; Fighting police; Mass property destruction; Thrown objects; Looting.
    \item \textbf{E) CAR ACCIDENT:} Vehicle-to-vehicle/pedestrian impacts; Flipping/rolling; High-speed crashes; Audio: Crash/screeching.
    \item \textbf{F) ABUSE:} Domestic/child abuse; Physical intimidation; Violent restraint; Aggressive domination; Elder abuse.
\end{itemize}
\textbf{NO ANOMALY = 0:} \textbf{Normal:} Walking, talking, shopping. \textbf{Sports:} Organized/gym/training (no harm). \textbf{Verbal:} Arguing (no contact). \textbf{Ambiguous:} Crowds/running (no violence).
}}
\vspace{-0.5em}
\caption{Exact anomaly scoring criteria for XD-Violence.}
\label{fig:xd_prompt}
\end{figure*}
\newpage
\begin{figure*}[h]
\centering
\setlength{\fboxsep}{3pt}
\fbox{\parbox{0.98\textwidth}{\scriptsize
\textbf{Ubnormal System Prompt Definitions:}\\[0.2em]
\textbf{STRICT RULES FOR ANOMALY SCORING} --- 1. \textbf{SCOPE:} Abuse, Arrest, Arson, Assault, RoadAccidents, Shooting, Shoplifting, Stealing, Vandalism, Fighting, Explosion. 2. \textbf{FACT-BASED:} Analyze ONLY explicit visuals. Do NOT assume intent. Consider duration/obstruction/risk cues.\\[0.2em]
\textbf{3. ANOMALY = 1 (Immediate or Emerging Public Safety Risk)}
\begin{itemize} \setlength{\itemsep}{0pt} \setlength{\parskip}{0pt} \setlength{\parsep}{0pt}
    \item \textbf{A) HUMAN CRIMINALITY:} Assault/Fighting (punches, kicks); Shooting (weapons, panic); Stealing/Burglary (concealment, force); Vandalism.
    \item \textbf{B) HAZARDOUS SITUATIONS:} \textit{RoadAccidents:} Collision; Vehicle on sidewalk/grass; Blocking active lanes; Stopped in traffic; Visible damage. \textit{Explosion/Arson:} Flash/blast; Thick smoke from fixed location; Open flames; Fleeing smoke.
    \item \textbf{C) ELEVATED RISK DUE TO PERSISTENCE:} Standing/lingering near entrance; Repeated pacing/monitoring; Loitering in roadway causing traffic changes. (Emerging risks only).
\end{itemize}
\textbf{4. NO ANOMALY = 0:} Walking/standing briefly; Jogging; Cars at red lights/congestion; Parked legally; Light fog/rain. If persistence/danger not explicit, classify as 0.
}}
\vspace{-0.5em}
\caption{Exact anomaly scoring criteria for UBNormal.}
\label{fig:ubnormal_prompt}
\end{figure*}

\subsection{Hyperparameters and Configuration}
\label{subsec:hyperparams}

Table~\ref{tab:hyperparams} summarizes the key hyperparameters used in our experiments.

\subsection{Parallel Processing Architecture}
\label{subsec:parallel}

To ensure data integrity during parallel processing, we enforce strict isolation between video processing threads:

\begin{enumerate}
    \item \textbf{Vector Memory Isolation:} Each video receives its own \texttt{VectorMemory} instance, preventing cross-video context contamination.
    \item \textbf{Shared Encoder:} The sentence embedding model is loaded once and shared across threads (read-only during inference), reducing memory overhead.
    \item \textbf{Model Locks:} Thread-safe locks (\texttt{vlm\_lock}, \texttt{llm\_lock}) serialize access to the VLM and LLM during generation.
\end{enumerate}

This architecture enables efficient parallel processing of multiple videos while guaranteeing zero cross-video data leakage, which is critical for accurate anomaly detection in surveillance applications.

\begin{table}[h]
\centering
\caption{QVAD Hyperparameters}
\label{tab:hyperparams}
\begin{tabular}{lcc}
\toprule
\textbf{Parameter} & \textbf{Symbol} & \textbf{Value} \\
\midrule
\multicolumn{3}{l}{\textit{Frame Selection}} \\
Uniform pre-sample count & -- & 32 \\
Motion-aware target count & $T$ & 8 \\
Gaussian blur kernel size & $\sigma$ & 21 \\
Motion threshold & $\delta$ & 25 \\
\midrule
\multicolumn{3}{l}{\textit{Token Limits}} \\
Max initial caption tokens & -- & 300 \\
Max VLM answer tokens & -- & 150 \\
Max memory context tokens & $\tau_{\text{ctx}}$ & 512 \\
Max enriched caption tokens & $\tau_{\max}$ & 2048 \\
Max stored caption tokens & $\tau_{\text{cap}}$ & 150 \\
\midrule
\multicolumn{3}{l}{\textit{Conversation Control}} \\
Max conversation turns & $K$ & 2 \\
Confidence threshold & $\theta$ & 0.7 \\
\midrule
\multicolumn{3}{l}{\textit{Vector Memory}} \\
Embedding dimension & -- & 384 \\
Max memory size & $M_{\max}$ & 400 \\
Retrieval count & $k$ & 3 \\
\midrule
\multicolumn{3}{l}{\textit{Model Generation}} \\
VLM max new tokens & -- & 512 \\
VLM temperature & -- & 0.7 \\
LLM max new tokens & -- & 512 \\
LLM temperature & -- & 0.3 \\
\bottomrule
\end{tabular}
\end{table}

\begin{table}[t]
\centering
\caption{Optimal values for hyper-parameters: $\sigma_1$, $\sigma_2$ and $\alpha$}
\label{tab:post-process hyperparameters}
\begin{tabular}{c|c|c|c|c|}
\cline{2-5}
 & UCF-Crime & UB-Normal & XD-Violence & Complex-VAD \\ \hline
\multicolumn{1}{|c|}{$\sigma_1$} & 280 & 145 & 142 & 420 \\ \hline
\multicolumn{1}{|c|}{$\sigma_2$} & 0.3 & 0.5 & 0.7 & 0.4 \\ \hline
\multicolumn{1}{|c|}{$\alpha$} & 0.05 & 0.2 & 0.4 & 0.21 \\ \hline
\end{tabular}
\end{table}
\subsection{Optimizing Post-Processing Parameters}
The hyperparameters $\sigma_1$, $\sigma_2$, and $\alpha$ are optimized on the validation set of each dataset (Table~\ref{tab:post-process hyperparameters}). Larger values of $\sigma_1$ are favored for datasets with longer and more complex anomaly durations, indicating the benefit of stronger temporal aggregation, while smaller values preserve sharper transitions for short-lived anomalies. The calibration parameter $\alpha$ used for anomaly score weighting in Eq. \eqref{eq:superadd} adapts to dataset-specific confidence distributions, mitigating over-suppression of anomalous frames and over-amplification of normal ones. Overall, these optimized parameters enable robust yet flexible temporal post-processing across diverse benchmarks.

\subsection{Implementation of VADTree on UBNormal and ComplexVAD}
\label{sub: implenVad}
We first pre-process the UBNormal and CVAD dataset into a unified video directory with associated track annotation files. Temporal boundaries are extracted using EfficientGEBD with a ResNet-18 backbone, initialized from Kinetics-GEBD and fine-tuned on UBNormal. The model predicts per-frame boundary confidence scores in 100-frame windows and a threshold at 0.5, to form boundary timestamps and grouped into scene segments.

These segments are organized into a Hierarchical Graph Tree (HGTree) with coarse nodes capturing long-duration activities and fine nodes modeling short transitions. Redundant nodes are removed, and parent–child relations are built via temporal containment. For each node, motion-aware frame selection produces a compact visual summary by retaining the first and last frames and selecting high-motion frames based on grayscale differencing with Gaussian smoothing.

Selected frames are then processed by a VLM (LLaVA-Video-7B-Qwen2) using a static prompt to generate semantic scene descriptions and preliminary anomaly judgments, guided by UBNormal specific priors. The resulting captions are then scored by an LLM (DeepSeek-R1-Distill-Qwen-14B), which assigns anomaly values in [0,1] through structured reasoning.

Finally, node-level scores are refined via intra-cluster similarity aggregation using ImageBind features and fused across coarse and fine hierarchies to produce frame-level anomaly predictions.

\section{Baseline Methods}
\label{sec:appendix_baselines}

We compare \textbf{QVAD} with a diverse set of representative video anomaly detection (VAD) baselines, covering both \emph{specialized} and \emph{training-free} paradigms. All reported results are taken from the original papers or reproduced using the authors' released code and recommended evaluation protocols.

\subsection{Specialized Methods}

\textbf{HL-Net}~\cite{wu2020multimodal}

\textbf{RTFM}~\cite{tian2021rtfm}

\textbf{VadCLIP}~\cite{wu2024vadclip}

\textbf{VERA}~\cite{ye2025vera}

\textbf{Holmes-VAU}~\cite{zhang2025holmes}

\textbf{HyCoVAD}~\cite{hemmatyar2025hycovad}

\textbf{MLLM-EVAD}~\cite{mumcu2025leveraging}

\subsection{Training-free Methods}

\textbf{LAVAD}~\cite{zanella2024lavad}

\textbf{EventVAD}~\cite{shao2025eventvad}

\textbf{PANDA}~\cite{yang2025panda}

\textbf{VADTree}~\cite{li2025vadtree}

\textbf{Follow the Rules}~\cite{yang2024rules}
\end{document}